\documentclass[pdflatex,sn-mathphys-num]{sn-jnl}

\usepackage{graphicx}%
\graphicspath{{./}} 
\usepackage{multirow}%
\usepackage{amsmath,amssymb,amsfonts}%
\usepackage{amsthm}%
\usepackage{mathrsfs}%
\usepackage[title]{appendix}%
\usepackage[normalem]{ulem}
\usepackage{xcolor}%
\usepackage{textcomp}%
\usepackage{manyfoot}%
\usepackage{booktabs}%
\usepackage{algorithm}%
\usepackage{algorithmicx}%
\usepackage{algpseudocode}%
\usepackage{listings}%
\usepackage{braket} %

\theoremstyle{thmstyleone}%

\theoremstyle{thmstyletwo}%

\theoremstyle{thmstylethree}%

\begin{document}

\title[Article Title]{Quantum Machine Learning via Contrastive Training}

\author*[1,2,3]{Liudmila A. Zhukas}\email{liudmila.zhukas@duke.edu}
\equalcont{These authors contributed equally to this work.}

\author*[1,2,3]{Vivian Ni Zhang}\email{vivian.zhang@duke.edu}
\equalcont{These authors contributed equally to this work.}

\author[1]{Qiang Miao}
\author[4]{Qingfeng Wang}
\author[1,2,3]{Marko Cetina}
\author[1,2,3]{Jungsang Kim}
\author[3]{Lawrence Carin}
\author[1,2,3]{Christopher Monroe}

\affil[1]{Duke Quantum Center, Duke University, Durham, NC 27701}
\affil[2]{Department of Physics, Duke University, Durham, NC 27708}
\affil[3]{Department of Electrical and Computer Engineering, Duke University, Durham, NC 27708}
\affil[4]{Department of Physics and Astronomy, Tufts University, Medford, MA 02115}

\abstract{Quantum machine learning (QML) has attracted growing interest with the rapid parallel advances in large‑scale classical machine learning and quantum technologies. Similar to classical machine learning, QML models also face challenges arising from the scarcity of labeled data, particularly as their scale and complexity increase.
Here, we introduce self-supervised pretraining of quantum representations that reduces reliance on labeled data by learning invariances from unlabeled examples. We implement this paradigm on a programmable trapped-ion quantum computer, encoding images as quantum states. \textit{In situ} contrastive pretraining on hardware yields a representation that, when fine-tuned, classifies image families with higher mean test accuracy and lower run-to-run variability than models trained from random initialization. Performance improvement is especially significant in regimes with limited labeled training data. We show that the learned invariances generalize beyond the pretraining image samples. Unlike prior work, our pipeline derives similarity from measured quantum overlaps and executes all training and classification stages on hardware. These results establish a label-efficient route to quantum representation learning, with direct relevance to quantum-native datasets and a clear path to larger classical inputs.}

\maketitle

\section{Introduction}\label{sec1}

Throughout most of its development \cite{bishop,hastie,murphy}, machine learning has assumed access to large quantities of {\em labeled} data. In many settings, acquiring {\em labeled} data is difficult, because obtaining the associated labels often requires human annotation or interpretation. The same problem arises, for example, in computational chemistry, where obtaining labels of molecular properties requires complicated and time-consuming lab experiments~\citep{Wang2022}. By contrast, it is increasingly possible to access large quantities of {\em unlabeled} data. As an example, while it is difficult to get images that have been carefully annotated and labeled, there are vast quantities of unlabeled images available on the web and from other sources. Training models with unlabeled data is often termed ``unsupervised learning.''

So motivated, there has been recent interest within machine learning research to {\em pretrain} models with unlabeled data, of which there often is a vast quantity, and then to {\em fine-tune} these models subsequently with a typically relatively small quantity of labeled data. Ideally, if pretraining is done well, a high-quality final model may be realized even with relatively sparse labeled data. There are many recent examples of such pretraining with unlabeled data, among the most prominent being language \cite{GPT2} and vision models \cite{ViT-MAE}.

In this paper, we consider a different class of unsupervised pretraining, termed contrastive learning \cite{contrastivelearning}. As we detail in Section~\ref{sec2}, contrastive learning augments a set of unlabeled data (often images) in a manner that allows unsupervised training via classical methods that assume access to labeled data.

Most work on contrastive learning utilizes classical computing systems \cite{contrastive_visual,NEURIPS2020_d89a66c7}; however, the rapid advance of quantum technology offers a new route to implementing contrastive learning methods on quantum computing systems. A prior study shows that a contrastive supervised learning pipeline implemented with quantum circuits can achieve high accuracy on handwritten-digit images using 25 labeled examples per class. However, in this work, the images were reduced to $8\times 8$ pixels, and with the contrastive learning performed in simulation, where only the final classification stage is demonstrated on a quantum processor~\cite{QSupCon}. Our work differs from \cite{QSupCon} in three key ways: ($i$) the entire contrastive learning pipeline is executed on quantum hardware; ($ii$) pairwise similarity for the contrastive objective is computed on hardware via a quantum state-overlap measurement, whereas in \cite{QSupCon} the similarity is calculated as a classical inner product in a projected feature space; ($iii$) the image augmentations we consider were implemented in the domain of the original image, e.g., rotation invariance. In \cite{QSupCon}, the augmentations were performed in the quantum domain, which may not necessarily track to the desired feature invariance to augmentations in the original images (quantum augmentations may be more appropriate for data that are originally quantum).

In this article, we present a quantum contrastive learning approach that captures essential similarities and differences within the quantum representation of classical datasets. To realize this, we encode classical images as quantum circuits and define a contrastive loss using a similarity score measured on a quantum processor. The method follows a two-stage protocol: (1) self-supervised representation learning on unlabeled data (pretraining) and (2) supervised classifier training on a small labeled set (fine-tuning), enabling efficient contrastive learning within the quantum domain. This two-stage technique aligns with recent work in classical machine learning \cite{GPT2,ViT-MAE}, reducing the quantity of labeled data needed, and it improves generalization to previously unseen classes or variations. 

We describe how both stages of this protocol are realized on a universal trapped-ion quantum processor, where the variational circuits are parametrized and executed with real-time feedback. We find that the proposed method significantly outperforms training in a supervised manner on the labeled data (without pretraining); it shows higher accuracy and stability across all numbers of training samples. To our knowledge, this is the first experimental demonstration of \textit{in situ} contrastive self-supervised pretraining on the universal trapped-ion quantum processor, followed by downstream fine-tuning on the same hardware. \par

\section{Setup}\label{sec2}

\begin{figure}[h]
\centering
\includegraphics[width=\textwidth]{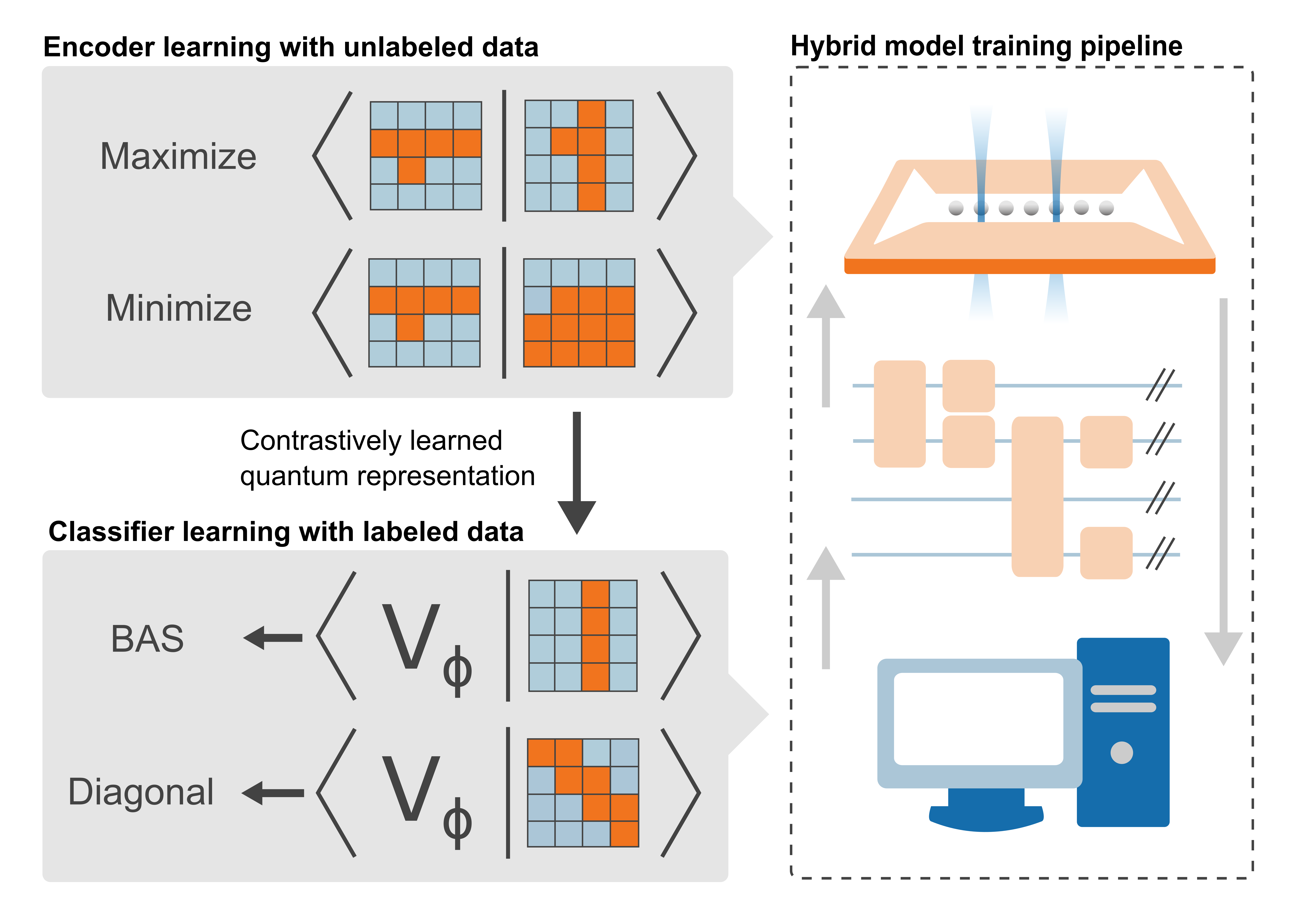}
\caption{\textbf{Overview of the contrastive learning workflow implemented on the trapped-ion quantum processor.} First, we encode each 4 × 4 binary image as a parameterized four-qubit circuit, where the rotation angles represent pixel patterns. When each image is mapped to a circuit, we learn a quantum representation that captures essential differences and similarities of selected classes. For this, each pair of unlabeled images is compared directly on hardware by executing a single circuit, and the results of such comparison over all pairs are aggregated on the classical optimizer to compute the contrastive loss. During each step, the classical optimizer updates the gate angles in the shared representation. By training the model with a chosen circuit structure, we are able to discover a representation where images from the same class cluster together even when they differ by rotation, while images from different classes are pushed apart. Finally, we evaluate the test accuracy on the quantum processor.}
\label{fig:main}
\end{figure}

\paragraph{Hardware} 
We conduct the experiment using four qubits selected from a seven-ion programmable trapped-ion quantum computer, featuring all-to-all connectivity, individual optical addressing, and state readout for all qubits~\cite{egan_BS,Zhukas2024} (see Appendix for more description). Specifically, we work with hyperfine qubits encoded in $^{171}$Yb$^{+}$ ions, which are confined in a linear chain in a micro-fabricated surface-electrode trap. The ions are individually addressed using tightly focused laser beams that couple qubits to the shared motional modes of the ions, see Fig.~\ref{fig:main}. This configuration enables the execution of universal gates on any chosen individual qubit or arbitrary qubit pair. Native two-qubit operations are realized via M{\o}lmer–S{\o}rensen interactions between selected qubit pairs~\cite{Sorensen:99}. In this work, we achieve two-qubit gate fidelities of approximately 99\% and a single-qubit gate fidelity of approximately 99.5\% (excluding state preparation and measurement [SPAM] errors, which are estimated to be around 0.4\%).

\paragraph{Contrastive Learning} 

The first application of contrastive learning dates to \cite{hinton_contrastive}. As summarized below, we employ the relatively simple modern approach developed in \cite{contrastivelearning}. In the pretraining phase,  {\em unlabeled} data are employed to develop a model that is effective at extracting {\em features} from a class of data, with the goal that these features are useful for general-purpose classification tasks. Specifically, let $f_\theta(x):\mathbb{R}^n\mapsto\mathbb{R}^d$ represent a function, with parameters $\theta\in\mathbb{R}^m$, where $d$ represents the dimension of the features (in the quantum implementation, the features are complex, but for introducing contrastive learning, we consider real quantities). The parameters $\theta$ are first learned in the pretraining phase, to perform a learning task based on the unlabeled data. The feature-representation model $f_\theta(x)$ is then {\em reused} for the specific classification problem of interest, using a typically small set of labeled data. In the fine-tuning phase, the parameters $\theta$ learned during pretraining serve as the initial feature-model parameters, which are then refined (fine-tuned) using labeled data from the supervised learning task.

\begin{figure}[h]
\centering
\includegraphics[width=0.8\textwidth]{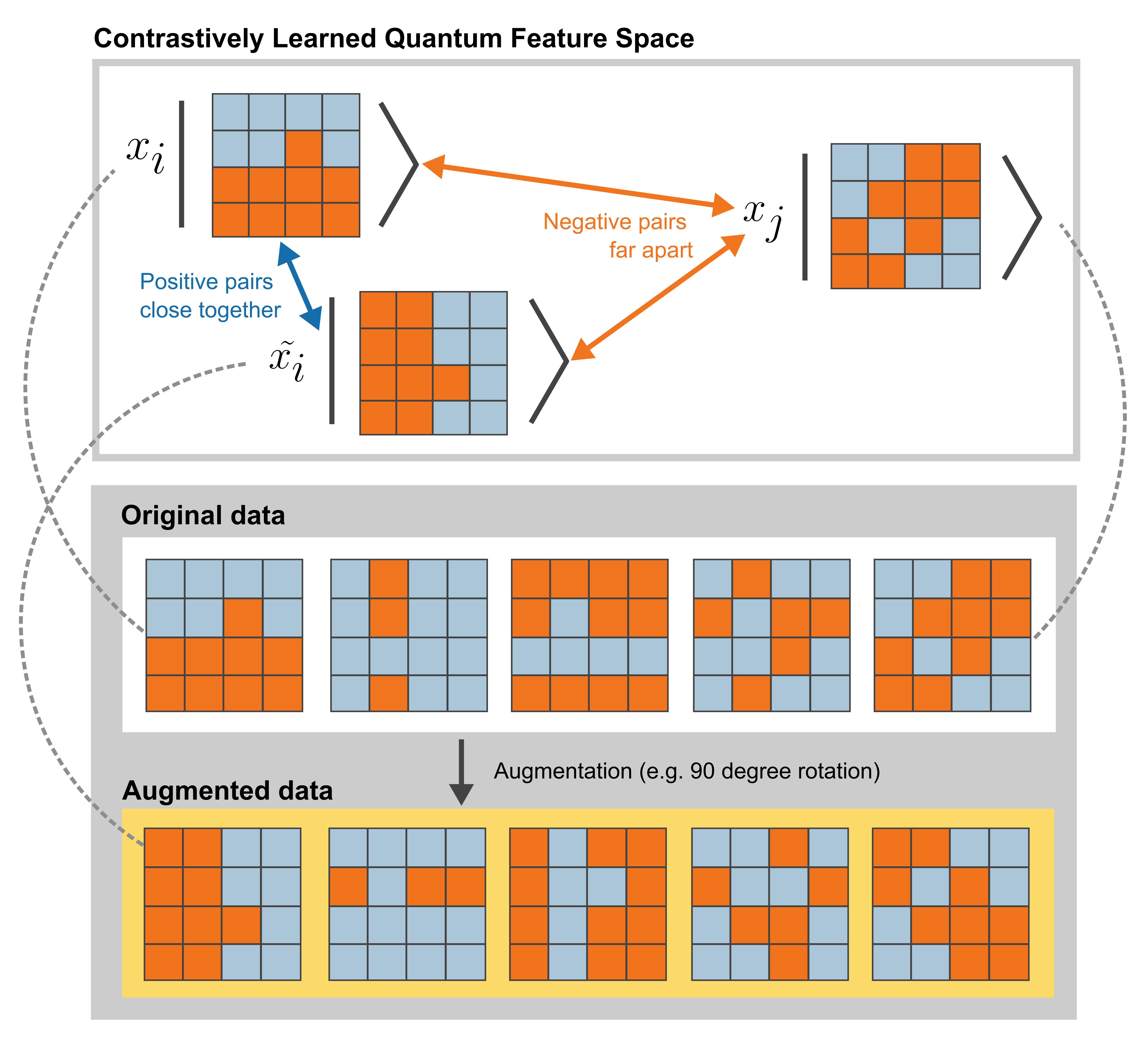}
\caption{\textbf{Constructing a contrastively learned quantum feature space for classical images}. We first take $N_U$ classical images, apply augmentation (such as 90-degree rotations) to each image to obtain an image set of $2N_U$. We then learn a quantum representation of these classical images, where augmented (distinct) images form positive (negative) example pairs that are similar (dissimilar) in the representation space. This process does not rely on data labels. The image set shown here ($N_U = 5$) contains binary images where orange (1) and blue (0) indicate the binary pixel values.}
\label{fig:contrastive}
\end{figure}

Consider a set of {\em unlabeled} images $x_i$, for $i=1,\dots,N_U$ (we assume each $x_i$ is an image, as that specifies the form of data-perturbations we consider, and our experiments consider such image data. The first step of contrastive learning is data-augmentation: Each $x_i$ is mapped to a new image $\tilde{x}_i$ through some augmentation, often done at random. Two common examples concern adding noise to the pixels of $x_i$, or rotating the image by an angle (which could be chosen at random). In Figure~\ref{fig:contrastive} we show an example $x_i$ and its corresponding $\tilde{x}_i$, of relevance to our experiment, discussed in detail later. After this data-augmentation phase, we now have $2N_U$ unlabeled samples $(x_i,\tilde{x}_i)$ for $i=1,\dots,N_U$. In the subsequent discussion, we consider the total dataset $\{x_i\}$ for $i=1,\dots,2N_U$, where sample $x_i$ has the augmented version $x_{N_U+i}$.

We now set up a learning objective with the following goal: For $i=1,\dots,N_U$, the model $f_\theta(x_i)$ should yield features that are more similar to features $f_\theta(x_{N_U+i})$ than they are to $f_\theta(x_j)$ for $j\neq i$ and $j\neq N_U+i$. For this to work, the feature model must be invariant to the aforementioned augmentation, implicitly focusing on features of the image (before and after augmentation) that are similar. The idea of contrastive learning is that these features effectively capture the most descriptive aspects of an image, and are therefore effective for many image-classification tasks, including those for which we will do fine-tuning.

Letting $z_i=f_\theta(x_i)$, consider the following loss function:
\begin{equation}
    \ell_i^{(U)} (\theta)=\frac{\exp\!\left((z_i^\top z_{N_U+i})/\tau\right)}{\sum_{j=1}^{2N_U}(1-\delta_{i,j})(1-\delta_{N_U+i,j})\exp\!\left((z_i^\top z_{j})/\tau\right)}\label{eq:contrastive_loss}
\end{equation}
where $\delta_{i,j}$ is the Kronecker delta function, and $\tau$ is a ``temperature'' parameter that can be tuned. The learning objective for contrastive learning is
\begin{equation}
    \hat{\theta}_{\text{pre-train}}=\mbox{argmin}_\theta \left( -\sum_{i=1}^{N_U} \log \ell_i^{(U)}(\theta) \right)\label{eq:contrastive_loss_total}
\end{equation}
The contrastive-learning objective seeks features such that sample $i$ and its augmented version (sample $N_U+i$) look more similar than it does to all other samples and their augmented versions. 

Consider a {\em labeled} training set $(\chi_i,y_i)$ for $i=1,\dots,N_L$, where $\chi_i\in\mathbb{R}^n$ is the same {\em class} (form) of covariates as applied in the pretraining phase, but for a specific supervised-learning setting. 

For simplicity, we assume $y_i\in\{0,1\}$ (binary classification), and the loss function is
\begin{equation}\label{eq:sup_loss}
    \ell_i^{(L)}(\theta,\phi)= \left[S(w^\top z_i+b)\right]^{y_i}\!\left[1-S(w^\top z_i+b)\right]^{1-y_i},~~~S(\alpha)=\frac{e^\alpha}{1+e^\alpha}
\end{equation}
where $\theta$ are the same parameters associated with the feature extractor, and $\phi=(w,b)$ are additional parameters connected to the (generalized linear) classifier, where $w\in\mathbb{R}^d$ and $b\in\mathbb{R}$. Learning proceeds as
\begin{equation}
    (\hat{\theta}_{\text{fine-tuned}},\hat{\phi})=\mbox{argmin}_{(\theta,\phi )}\left(-\sum_{i=1}^{N_L}\log \ell_i^{(L)}(\theta,\phi)\right)
\end{equation}
where when training $\theta$ is initialized at $\hat{\theta}_{\text{pre-train}}$, while $\phi$ is initialized at random. After the model is so trained, for test data $\chi_{N_{\!L}+1}$, the probability that $y_{N_{\!L}+1}=1$ is $S[\hat{w}^\top f_{\hat{\theta}_{\text{fine-tuned}}}(\chi_{N_{\!L}+1})+\hat{b}]$.

\begin{figure}[h]
\centering
\includegraphics[width=\textwidth]{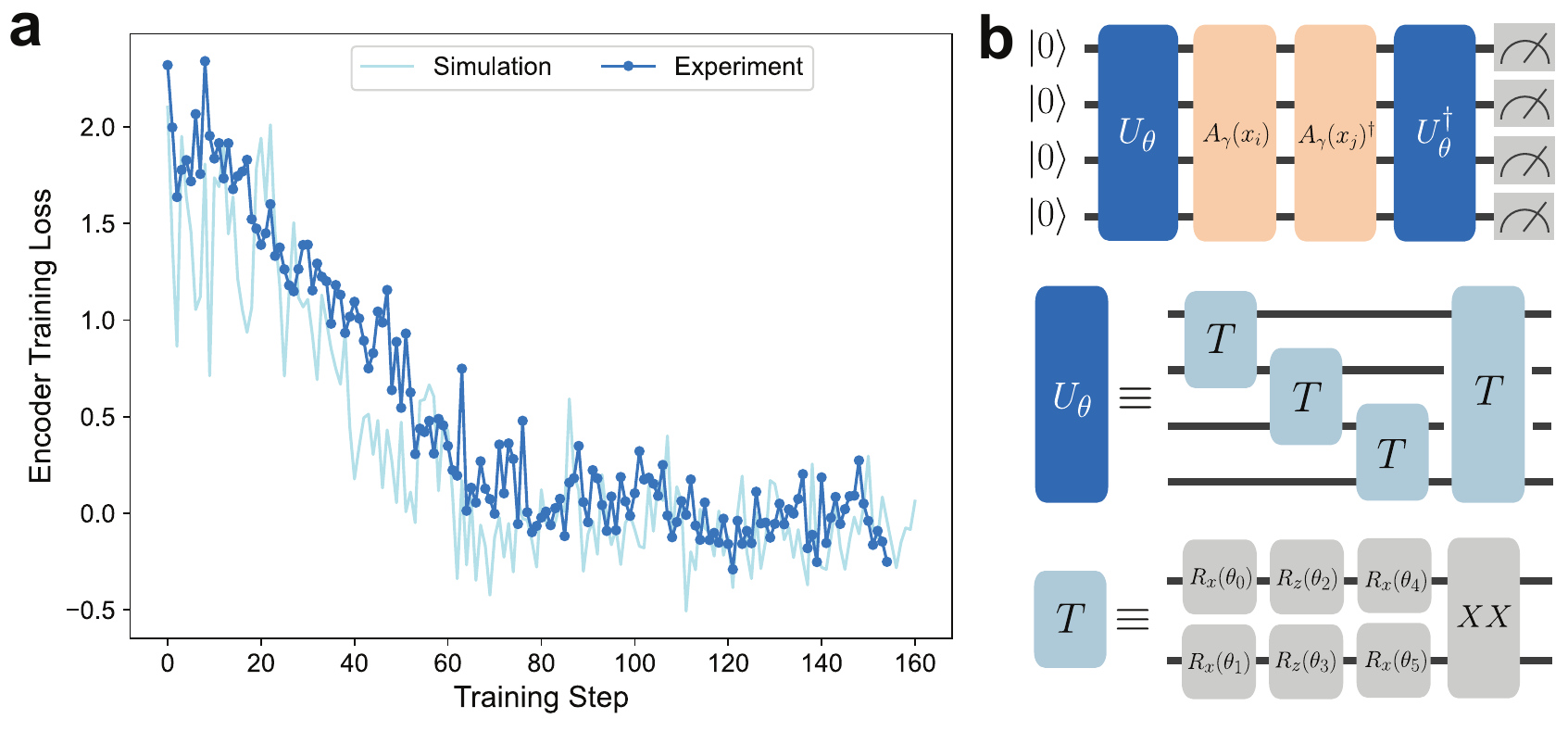}
\caption{\textbf{Encoder training results and circuit structure}. a) Encoder training loss. The light blue solid line represents a noiseless simulation using exact statevector evaluation. The blue curve shows experimental results on the quantum processor.  b) Circuit structure for the encoder training: the variational unitary $U_\theta$ and the data-encoding unitary $A_\gamma(x)$ are parameterized circuits acting on four quantum bits. $A_\gamma(x_i)$ and $A_\gamma(x_j)$ are learnable quantum encoding of classical images $x_i$ and $x_j$. Utilizing ``all-to-all" connectivity on the trapped-ion quantum processor, we impose operator \textbf{T} on qubit pairs, which contains alternating $\mathbf{R}_x$ and $\mathbf{R}_z$ single-qubit operations followed by a fixed-angle entangling gate. We encode variational parameters into the single-qubit operations while all two-qubit gate operations are performed with a fixed angle of $\pi/2$.}
\label{fig:encoder_and_circ}
\end{figure}

In the quantum setup, the feature map \(f_\theta(x)\) is realized by the state-preparation circuit \(A_{\boldsymbol\gamma}(x)U_{\boldsymbol\theta}\), so \(z_i\) is replaced by the normalized state \(\ket{\psi(x_i)}=A_{\boldsymbol\gamma}(x)U_{\boldsymbol\theta}\ket{0}\), and the similarity \(z_i^\top z_j\) in the contrastive objective is replaced by the squared overlap \(s(x_i,x_j)=\lvert\braket{\psi(x_i)\mid\psi(x_j)}\rvert^{2}\), measured with the state-overlap circuit shown in Fig~\ref{fig:encoder_and_circ}(b). The contrastive loss is then computed directly from these hardware-measured overlaps, and the parameters \((\boldsymbol\gamma,\boldsymbol\theta)\) are updated via SPSA~\cite{Spall, spall1998overview, tran2024variational}.

\paragraph{Image Dataset} 
We demonstrate the effectiveness of contrastive quantum feature learning and classification on $4\times4$ binary images taken from two distinct classes: horizontal or vertical bar patterns (commonly known as ``Bars-and-Stripes'', or BAS) and diagonal stripe patterns (see Fig.~\ref{fig:BAS_and_diagonal} in Appendix). 
These test datasets are visually interpretable and are widely used in quantum and classical benchmarks \cite{MacKay2003}. 
Each $4\times4$ binary image has a natural mapping onto a four‑qubit register, where the 16-bit classical data correspond directly to 16 computational bases of the Hilbert space. Utilizing ``all-to-all" connectivity of the trapped-ion processor and highly entangling circuit design, we can construct quantum circuits with variable rotation angles to encode any $4\times4$ binary pattern directly into the four-qubit register with minimal circuit depth.

\paragraph{Encoder architecture and self‑supervised objective}

The contrastively trained encoder produces a quantum representation of the input images, where each example and its augmented version are similar, while unrelated examples remain dissimilar. The augmentation in this case is a $90^{\circ}$ clockwise rotation of the image. By choosing perturbations that reflect this desired label invariance, the contrastive training learns a representation that is robust to such variations and is effective for downstream classification.

To demonstrate that the robustness of the encoder against augmentation is generalizable beyond the two families of images used in classification, we train and evaluate the encoder model using examples outside of the classification problem space. The encoder’s training and test datasets consist of single-pixel perturbations of the canonical shapes, see Fig.~\ref{fig:contrastive}.
This choice of encoder training images is intended to emulate real-world use cases of contrastive learning involving natural images. In particular, the examples used for contrastive training need not be identical to those encountered in downstream tasks; however, they should be drawn from a similar underlying distribution (e.g., a distribution of meaningful real-world images).

In the encoder training dataset, $N_U$ $4\times4$ images are sampled from the perturbed images mentioned above and paired with their augmented versions to form a positive instance. Cross-family pairings act as negatives, forming a training dataset of size $2N_U$. The encoder training is self-supervised without the use of labels.

In the encoder training set, each classical image $x \in \{0,1\}^{16}$ is mapped to a four-qubit quantum state.
\begin{equation}
  \bigl|\psi(x)\bigr\rangle = A_{\boldsymbol\gamma}(x)U_{\boldsymbol\theta}\,\,\lvert 0\rangle^{\otimes 4}.
\end{equation}

The data-encoding unitary $A_{\boldsymbol\gamma}(x)$, which is a quantum circuit parametrized by the angle vector $\gamma$ of size 4, encodes the image $x$ into a quantum representation. It first repeats the same encoding circuit on the four quadrants of the image consecutively, and finally on an aggregation of the four quadrants, see Fig.~\ref{fig:Ax} in the Appendix for more details.
The variational unitary block $U_{\boldsymbol\theta}$ alternates four fixed XX gates with six $\mathbf{R}_z\mathbf{R}_x\mathbf{R}_z$ triples per pair of qubits, yielding 24 tunable parameters $\theta$, see Fig.~\ref{fig:encoder_and_circ}(b).
\par

The similarity between two images is evaluated as the inner product of the corresponding quantum features.
\begin{equation}
  s(x_i,x_j) = \bigl|\langle\psi(x_i)\mid\psi(x_j)\rangle\bigr|^{2},\label{eq:sim}
\end{equation}
which we compute on the quantum processor. 
Namely, the similarities are measured via the state‐overlap circuits $\bra{0}U_\theta^\dag A_\gamma(x_{i})^\dag A_\gamma(x_{j}) U_\theta\ket{0}$ for $i, j \in [0,2N_U)$, where the circuit component that encodes $x_j$ is performed as is while that of $x_i$ is applied in its inverse (adjoint) order  (see Fig.~\ref{fig:contrastive}). Recalling the contrastive-learning setup in \eqref{eq:contrastive_loss}, the inner product there as $z_i^\top z_j$ is now replaced by the quantum version in \eqref{eq:sim}. In practice, we use $\tau = 0.2$, the parameter we determine empirically from the simulations. 

\paragraph{Classifier fine‑tuning and evaluation}

Considering the classifier learned via \eqref{eq:sup_loss}, we here assume $b=0$, and we discuss the conversion of inner product $w^\top f_\theta(x)$ to its quantum equivalent. After the encoder learns a quantum representation robust against rotation, we train a quantum classifier to distinguish between the two image classes. The learned feature encoding circuit $A_{\boldsymbol\gamma}(x)U_{\boldsymbol\theta}$ is combined in sequence with another parameterized circuit $V_{\boldsymbol\phi}$ with 18 parameters. We evaluate $\bra{0} V_{\boldsymbol\phi}A_{\boldsymbol\gamma}(x)U_{\boldsymbol\theta}\ket{0}$ for all examples $x$ in a balanced labeled training data set randomly drawn from the BAS and diagonal images, and update the model using binary cross-entropy loss. $V_{\boldsymbol\phi}$ has a circuit structure similar to $U_{\boldsymbol\theta}$ as presented in Figure~\ref{fig:encoder_and_circ}(b), except with one less module $T$ connecting qubit 0 and 3.

We evaluate the performance of two classifier training regimes. The \emph{encoder‑assisted training} warm‑starts $\theta$ from the encoder results and fine‑tunes it jointly with the learning of $\phi$ while keeping $\gamma$ fixed (See Appendix for details on the fine-tuning procedure).
The other regime is the \emph{randomly initialized} control, where the initial parameters in $\gamma$, $\theta$, and $\phi$ are initialized with random values; however, the circuit structure and the number of tunable parameters are preserved. 
We allow all parameters to be updated at the same learning rate in this setting.

\section{Results}\label{results}

\paragraph{Encoder Training}
The evolution of contrastive loss $\ell^{(U)}$ during encoder training is shown in Fig.~\ref{fig:encoder_and_circ}(a). We train the encoder on the quantum processor until the loss gradient over the last 20 iterations approaches zero, indicating convergence. See Fig.~\ref{fig:contrastive} for the $N_U= 5$ dataset used in training. The contrastive loss is computed as \eqref{eq:contrastive_loss_total} over all pairs of images in this dataset. The performance of the model is then evaluated by executing quantum circuits in all pairwise combinations of six unseen perturbed test examples and their augmented images. A measured expectation value below (above) 0.5 is interpreted as a negative (positive) prediction; the resulting accuracy is presented in Table~\ref{tab:train-test-acc}.

\begin{table}[t]
\centering
\caption{Encoder training and test accuracy using a decision threshold of 0.5\label{tab:train-test-acc}}
\begin{tabular}{lcc}
\hline
\textbf{Pair type} & \textbf{Training acc. (\%)} & \textbf{Test acc. (\%)} \\
\hline
Positive pairs & 100.0 & 85.0 \\
Negative pairs & 85.0  & 76.6 \\
Overall        & 86.7  & 72.3 \\
\hline
\end{tabular}
\end{table}

\paragraph{Classifier Training and Results}
The target image classes contain 28 BAS images and a total of 253 diagonal images. To obtain statistically robust performance estimates using noiseless simulations, we randomly partition the entire dataset into a training pool of $N_L$ labeled images, balanced between BAS and diagonal, and an independent test set of fixed size, also balanced across these two classes.
For each target training data size $N_L$, we repeatedly sample $N_L/2$ non-identical images from each class in the training pool. We classically simulate the training of two classifier models on this subset: one from random initialization and one by fine-tuning the contrastively pretrained encoder from the previous section. We then evaluate both models on the same test set, which contains 36 images unseen by the model during the training stage.

\begin{figure}[h]
\centering
\includegraphics[width=0.8\textwidth]{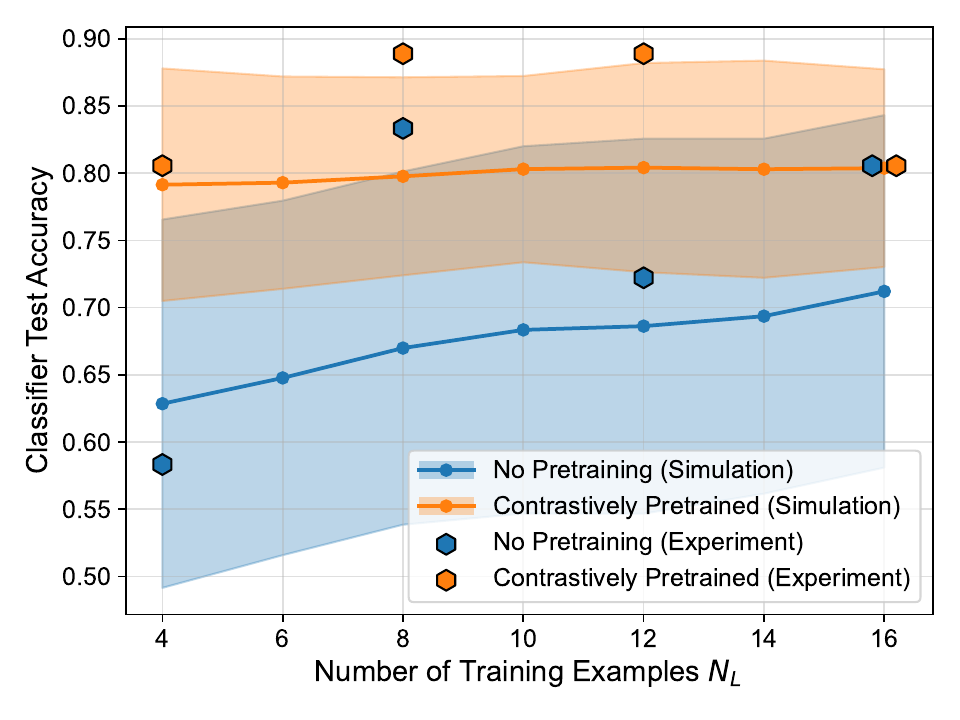}
\caption{\textbf{Classification stage results on the trapped-ion quantum processor.} We present classifier test accuracy as a function of the number of training examples. Solid blue and orange lines represent average simulated performance with and without contrastive pretraining, while the markers indicate experimental results on the quantum processor. Shaded regions represent $1\sigma$ standard deviation across 500 balanced train/test data partitions. In simulation, we sample multiple balanced training subsets for each dataset size, train both model types with identical circuit structure, and present the mean and standard deviation of test accuracy. A single dataset is used to evaluate both models on the quantum processor.}

\label{fig:results}
\end{figure}

Repeating this ``draw–train–test" cycle many times produces a distribution of accuracies; we report the mean and standard deviation of the distribution for each $N_L$ and each training regime, see Fig.~\ref{fig:results}. For the $N_L =$ 4, 8, 12, and 16, we sample one point from the data partition distribution above to verify on the quantum processor. \par 

Across all labeled training data budgets, the model assisted by a contrastively trained encoder provides both higher mean accuracy and lower variability than a network trained from random initialization.

The performance gap slightly closes as the benchmark model has more training data to learn the pattern from. The lower variance across train and test data selection is anticipated in the encoder-assisted method, as the pretraining initializes the model at a relatively optimized position. The experimental points follow the same trend. These results confirm that the representation learned in the self-supervised stage provides a robust starting point, whose advantage persists even as the amount of labeled data increases. 

We note that the contrastive models trained on the physical machine outperform the mean of simulations in most cases, possibly because of the selection of training data. With the random selection of training data in the simulation, oftentimes the training datasets contain augmented examples, which are considered redundant information for the classifier, as the encoder ensures that these examples are similar in the quantum feature space. In the models trained on the physical systems, we select the training data to avoid such redundancy. \par

We also note that at $N_L=8$, the benchmark model trained directly on the quantum hardware exhibits an accuracy higher than the mean of the simulation. Such out-of-distribution points arise because learning on the physical device corresponds to a single realization within the underlying distribution, as shown by the simulation results. Since the circuit expressiveness is identical, the group without pretraining may achieve comparable accuracy in a given run by chance. However, the simulated results indicate that pretraining yields more predictable performance and consistently higher accuracy on average.

During both simulation and training on the physical quantum computer, we limit the classifier training to 100 optimization steps, using 200 shots per circuit; simulations show that increasing this limit does not improve accuracy for either training regime. In Fig.~\ref{fig:results}, we use the same set of randomly chosen initial parameters for all training instances. We perform a separate check to assess robustness to various initialization parameters. In this test, simulations and experimental results show that the accuracy and stability advantage of the fine-tuned encoder over a randomly initialized model is consistent across different initial parameters, see~Fig.~\ref{fig:initialization_test}~(Appendix) for more details. \\

\section{Discussion}\label{discussion}

In this work, we implemented \textit{in-situ} contrastive self-supervised pretraining on a universal trapped-ion quantum processor applied to binary images. We pretrained the encoder on unlabeled data and then fine-tuned it for the downstream classification task.
Despite a limited shot budget and existing single- and two-qubit gate errors, this pipeline improves mean test accuracy and reduces variance relative to training identical quantum circuits from random initialization. It is worth noting that the encoder is trained on the perturbed images outside of the downstream label space, indicating that the learned representation captures the inherent invariance imposed by the augmentation (rotation) and can be generalized.

The absolute accuracy in this demonstration is limited by the protocol choices, which can be seen in the simulation results in Fig.~\ref{fig:results}. We note that the two classes are best separated in most simulation runs at a threshold different from 0.5, indicating a bias error in the classification model that can be addressed by adding more tunable parameters (see Appendix). The choice of an ansatz—a parameterized circuit architecture—for the unitaries $A_\gamma(x)$, $U_\theta$, and $V_\phi$ is primarily guided by empirical considerations based on a trade-off between circuit expressiveness and computational overhead.

The data encoder $A_\gamma(x)$ used in this work (Fig.~\ref{fig:Ax}) utilizes quantum circuits that are self-similar, to effectively capture self-similar features like BAS and diagonal images, while the training unitary $U_\theta$ and classification unitary $V_\phi$ are based on propagating entanglement locally at this small data sizes (Fig.~\ref{fig:encoder_and_circ}(b)). For larger image sizes, the choice of ansatz circuit structure should be optimized to capture the inherent patterns of the augmentation and classification, with the number of training parameters growing modestly with the image size.

\section{Outlook}\label{sec5}

To the best of our knowledge, we have performed the first experiment on a quantum computer using quantum contrastive learning to classify images. For this demonstration, we used $4\times4$ classical binary images and encoded the full image information into quantum states, although the same approach extends to grayscale images by mapping the grayscale value of a pixel to the coefficients of a quantum state vector. To scale up to larger datasets, a more efficient encoding scheme is needed. A promising approach is to first extract the most significant information from classical images and then encode this compressed information into the amplitudes of a quantum state. This can be achieved through methods like truncated singular value decomposition (SVD) or by drawing inspiration from well-established approximation methods in quantum physics, such as tensor network methods. For example, to produce an arbitrary $N$-qubit quantum state, one typically needs $\mathcal{O}(2^N)$ parameterized quantum gates. However, a tensor network state can compress the number of gates to a linear scale of $\mathcal{O}(N)$ with a controllable loss of accuracy. In Ref.~\cite{Iaconis2023}, the authors suggest an efficient amplitude-encoding scheme for classical images based on matrix-product states (MPS). This method scales as $N = \log_2 \textit{L}$ with the number of qubits $N$ for a $\textit{L}\times\textit{L}$ image, where $L$ is the number of pixels in one dimension. It is important to note that the circuit depth and two-qubit gate count scale logarithmically with image size as well. Considering this scaling, we estimate that training the encoder on  $16\times16$ grey-scale images and an 8-qubit register requires $\sim~60$ two-qubit entangling gates, where most of this budget comes from the encoding of the images and requires $\sim~50$ two-qubit entangling gates. This depth budget is compatible with present state-of-the-art quantum processors~\cite{PhysRevX.15.021052, Chen2024IonQ30, Kim2023}. With our circuit layout, we expect the linear growth of the number of trainable parameters with the qubit count for both training stages. The logarithmic growth in both qubits and depth makes scaling to larger inputs practical on industrial hardware, with a moderate increase in MPS bond dimension and trainable layers. It is possible that, as the problem size grows, the model trainability will be affected by the decay of the gradient magnitude \cite{Cerezo2021}. Similar to trends in classical machine learning, we expect that implementing larger models on hardware will drive methodological advances that improve the convergence behavior. For example, recent work suggests ``warm-start'' strategies that preserve the gradient signal early in optimization may improve convergence  \cite{10234243}. 

Beyond improving the classification of binary images, our work can be extended to processing quantum data. Unlike classical learning methods, which require collapsing quantum data into classical snapshots with limited samples in a stochastic manner, a native quantum learning approach can directly handle raw quantum data. This includes data generated by quantum computers, detected by quantum sensors, or teleported through quantum communication. Such a full-stack ``receive-process-learn" workflow on quantum hardware would fully leverage the native advantages of quantum computing and is poised to become a leading candidate for a flagship quantum application. Particularly, unlike other quantum machine learning methods, our \textit{in-situ} approach enables label-free learning, such as identifying quantum phases and their transitions in many-body systems. In this context, recent works point to these directions and motivate our approach. For example, theoretical studies have applied the contrastive method directly to quantum datasets for the unsupervised detection of out-of-distribution states \cite{PhysRevResearch.3.043184}. Semi-supervised frameworks map condensed matter measurements to feature vectors to classify phases on a classical simulator without explicit labels \cite{Han2023}.

\bmhead{Acknowledgements}

This work is supported by the DOE Quantum Systems Accelerator (DE-FOA-0002253) and the NSF STAQ Program (PHY-1818914).

\section{Appendix and Supplementary Material}\label{secA1}

\paragraph{Experimental System}

We perform all experiments on the trapped-ion quantum computer with a seven $^{171}$Yb$^{+}$ qubit register. In our system, each qubit can be individually addressed with a tightly focused laser beam, and arbitrary subsets of qubits can be targeted within a pulse sequence. A single focused beam provides independent amplitude, phase, and duration control for each ion. Two-qubit entangling gates are mediated by the collective motional modes of the ion chain, allowing native two-qubit gates between arbitrary pairs of ions. To make the two-qubit entangling gates robust to noise and motional mode drift, we construct segmented optical pulses~\cite{PhysRevA.97.062325}. At the end of the gate pulse sequence, we minimize residual coupling to the motion. Gate duration and optical pulse shaping are adjusted for each ion pair. A single-qubit gate represents a composite SK1 pulse~\cite{PhysRevA.70.052318}, where the amplitudes are calibrated for each qubit. We estimate single- and two-qubit gate fidelities and state-preparation-and-measurement (SPAM) errors as in Refs.~\cite{egan_BS, Zhukas2024}.

To characterize the noise of our trapped-ion system, we estimate hardware error sources using direct measurements and simulations of the hardware output. We report \textit{incoherent} two-qubit gate error rates in the $z$- and $x$-bases. For all ion pairs used in this work, the two-qubit gate error measured in the $z$-basis does not exceed 0.2$\%$, whereas an error in the $x$-basis does not exceed  0.1$\%$. The residual \textit{coherent} error is associated with the position (pointing) drift of the addressing laser beams relative to the ion chain and is corrected via periodic recalibration; a higher recalibration rate can further reduce its contribution. 

\paragraph{Hybrid Quantum-Classical Optimizations}

During the training of the models, each optimization step follows a hybrid quantum-classical scheme. In these hybrid models, classical data input and learnable parameters are encoded as tunable variables in parameterized quantum circuits, which are executed on quantum hardware and updated through a classical optimization routine. We employ automated calibration of single- and two-qubit gates, enabling the execution of a large number of circuits with periodic performance assessments. Specifically, we use a software-hardware API~\cite{wang2024cafqa} integrated with an automated calibration routine within the hardware control environment. This API transforms gate-level circuits into optimized laser pulse sequences and interfaces with the ARTIQ (Advanced Real-Time Infrastructure for Quantum Physics) control system~\cite{bourdeauducq_2016_51303}, which orchestrates pulse execution and triggers auto-calibrations as necessary. This software-hardware interface improves execution efficiency, supporting high-throughput circuit execution with minimal disruption. For the classical optimization of the variational quantum circuit (VQC), we choose the Simultaneous Perturbation Stochastic Approximation (SPSA) algorithm, a gradient-based method commonly used in VQE studies~\cite{Spall, spall1998overview, tran2024variational}.

\paragraph{Robustness to Randomly Chosen Initial Parameters}

\begin{figure}[h]
\centering
\includegraphics[width=0.8\textwidth]{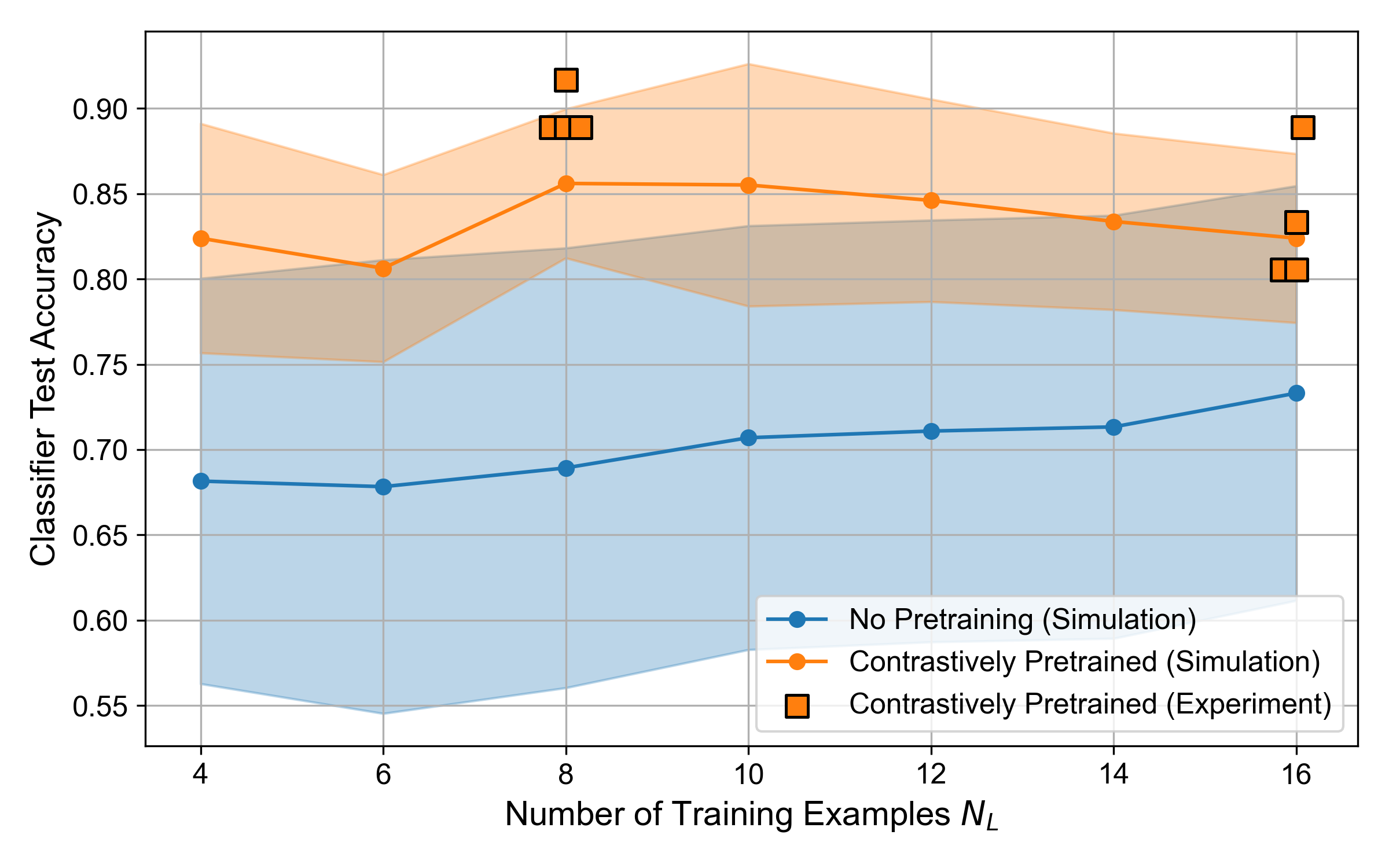}
\caption{\textbf{Robustness to initialization and the effect of pretraining.} The graph shows test accuracy versus the number of training examples in the classification stage. Solid lines represent average simulated performance with and without contrastive pretraining over 500 sets of randomly chosen initial parameters for each number of training examples; shaded regions indicate one standard deviation across all chosen parameter sets. Blue color represents the model without contrastive pretraining, while orange color represents a model initialized with a pretrained encoder. Orange squares show hardware results for four sets of randomly chosen initial parameters.}

\label{fig:initialization_test}
\end{figure}

We test optimization sensitivity to the choice of initial parameters. In Fig.~\ref{fig:initialization_test}, we show the effect of pretraining on models' robustness to the choice of such parameters. For each presented number of training examples, we repeat the training of the classifier model with randomly chosen initial parameters with (orange) and without (blue) contrastive pretraining. Solid curves represent the mean test accuracy over the 500 various sets of randomly selected initial parameters, and shaded regions denote $1\sigma$ standard deviation across all initializations. To test on hardware, we fine-tune the pretrained model on the trapped-ion processor using four randomly chosen initial parameter sets (orange squares). Pretraining consistently raises the mean accuracy and reduces variance across choices of initializations, indicating improved robustness; the experimental results follow the trend.

\paragraph{Training Examples}

We train the encoder and evaluate its performance using examples outside of the classification problem space. As shown in Fig.~\ref{fig:contrastive}, we sample $N_U$ images that differ by one pixel from the bars-and-stripes (BAS) and diagonal patterns, and then augment each image to create a training image set of size $2N_U$. We then take all pairs of images from the training image set and evaluate their similarities on the quantum processor. Pairs composed of rotated (unrelated) images are positive (negative) examples whose similarities the encoder seeks to maximize (minimize). The 10 images shown in Fig.~\ref{fig:contrastive} are the exact training set we use to train the encoder on the quantum hardware.

\begin{figure}[h]
\centering
\includegraphics[width=0.8\textwidth]{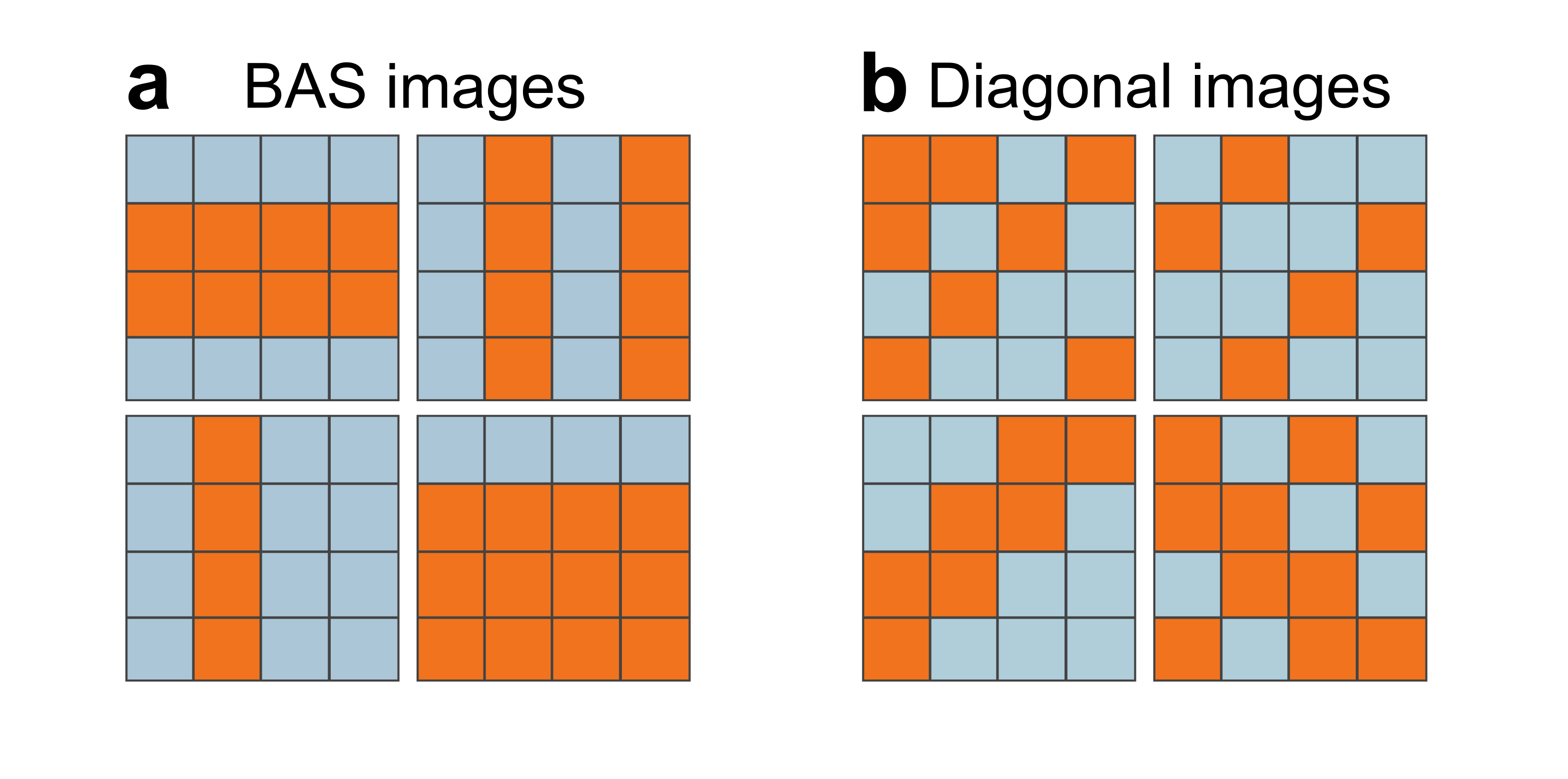}
\caption{\textbf{Example of an image set for the classification stage}. a) Images with all horizontal stripes or all bars. b) Images with diagonal stripes. These two classes are invariant to 90-degree rotation.}

\label{fig:BAS_and_diagonal}
\end{figure}

During the classification stage, the model distinguishes $4\times4$ binary images consisting of all horizontal stripes or all bars from images consisting of all diagonal stripes, as illustrated in Fig.~\ref{fig:BAS_and_diagonal}(a,b).

The encoder $A_\gamma(x)$ is a crucial component in the parametrized quantum circuit because the encoding method must preserve the property of the augmentation invariance in the original image in the quantum space. During the design of $A_\gamma(x)$, we followed the guideline that the augmentation of the classical data (e.g., reordering of $x$ that corresponds to a 90-degree rotation) should roughly map to an augmentation of the quantum data (e.g., a unitary transformation of the quantum feature). $A_\gamma(x)$ also needs to capture enough information from the original images in order to learn the difference between classes.

\begin{figure}[h]
\centering
\includegraphics[width=\textwidth]{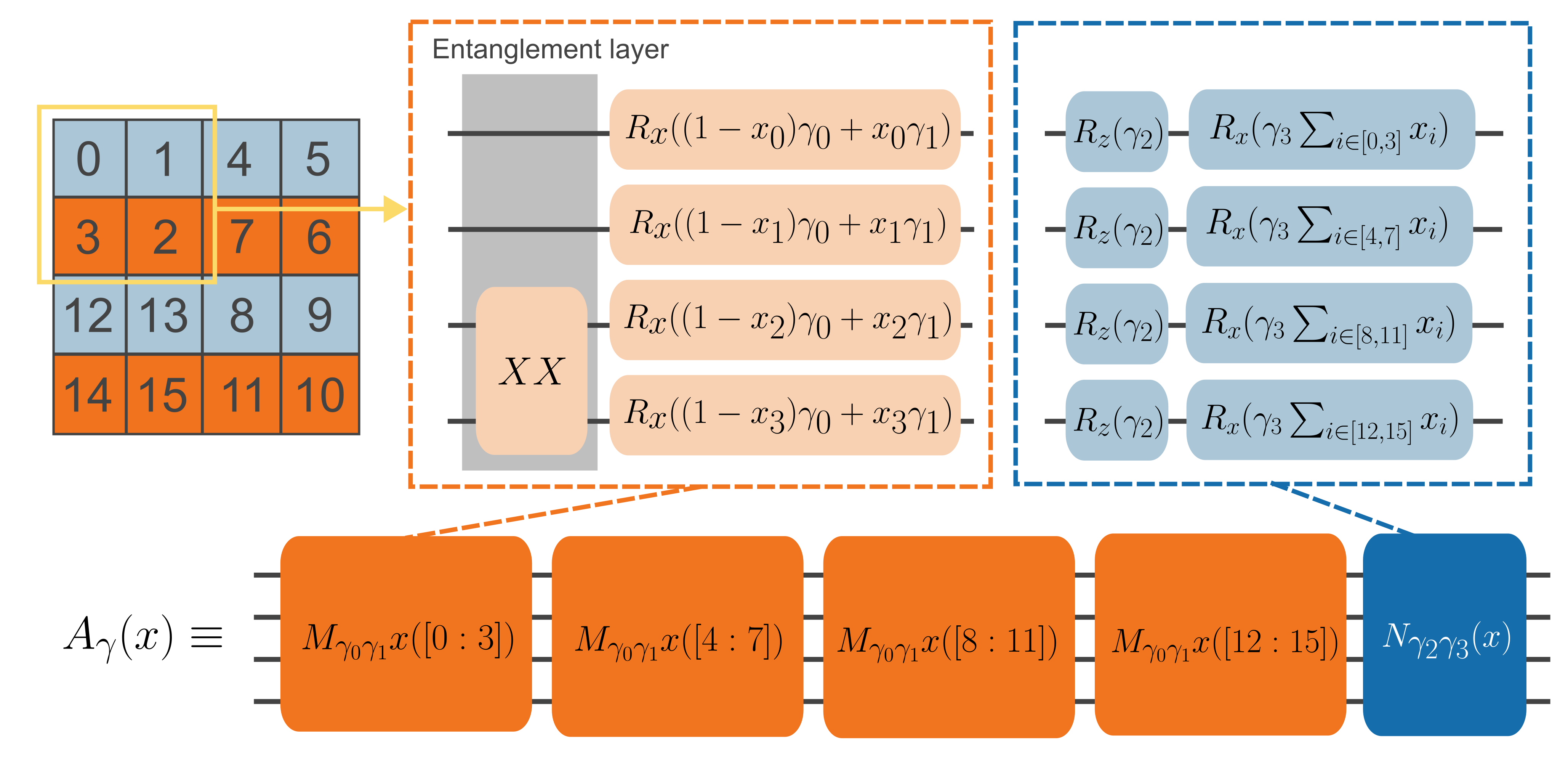}
\caption{\textbf{Circuit structure of $\mathbf{A_\gamma(x)}$}. We encode a $4\times4$ binary image $x$ into a four-qubit quantum register using the circuit $A_\gamma{(x)}$, where $\gamma$ is a vector of learnable angles of size four. The image on the top left shows how we flatten the binary image into a list of 16 binary numbers. This binary string is then segmented into four pieces, which are used as input to the same encoding module $M$. The entanglement layer in $M$ applies XX gates on pairs of qubits corresponding to binary value one (represented by orange). In the end, another encoding module $N$ acts on the sum of each segment.}
\label{fig:Ax}
\end{figure}

The ansatz circuit of $A_\gamma(x)$ of our choice is shown in Fig.~\ref{fig:Ax}. We segment the image into four quadrants and repeat the same encoding module on them in a convolutional manner. The encoding module contains an entangling layer where XX gates are placed between qubits corresponding to pixels valued one. For example, if positions 1 and 3 are both ones in a given quadrant while 2 and 4 are zeros, then the entangling layer contains a single XX gate between qubit 1 and qubit 3. The rest of the encoding module encodes each bit into a single qubit rotation on the corresponding qubit. The exact angles to which ones or zeros correspond are learned during the contrastive training stage (e.g., parameter $\gamma_0$ and $\gamma_1$). In the end, a separate module acts on an aggregation of the four quadrants---we sum each quadrant up and use the four aggregated numbers as the encoding input in this module. 

In practice, we separate the learning of the data-encoding unitary $A_\gamma(x)$ from that of the variational unitary $U_\theta$. The data-encoding unitary was optimized in a simulation setting where parameters $\gamma$ and $\theta$ are co-optimised \emph{in‑silico} on $N_U = 6$ images taken from the perturbed image distribution; only four parameters in $\gamma$ are retained. With the data-encoding unitary $A_\gamma(x)$ pretrained and fixed, the 24 angles in $\theta$ are then refined on the quantum processor from scratch. The learning of $U_\theta$ with the hybrid classical-quantum optimization routine uses a separate $N_U = 5$ dataset (see Fig.~\ref{fig:contrastive} for the exact dataset).

\begin{figure}[h]
\centering
\includegraphics[width=\textwidth]{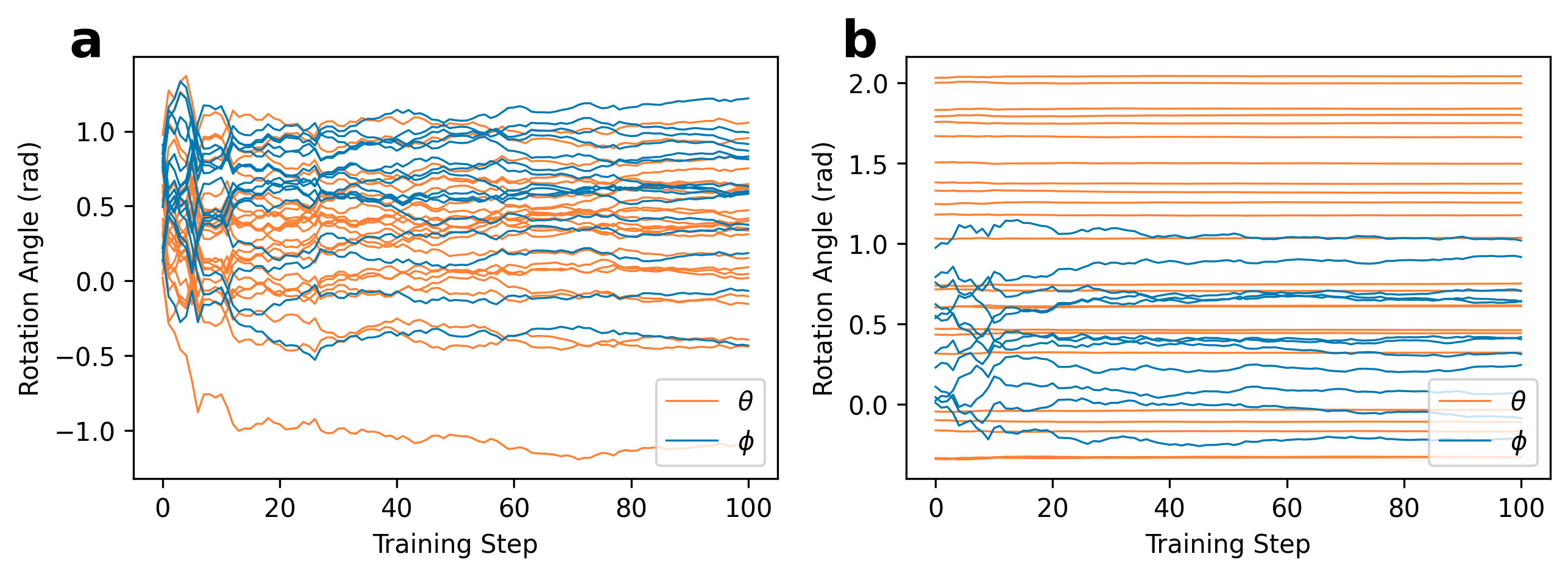}
\caption{\textbf{Evolution of learnable parameters during classifier training on the quantum hardware} a) In the randomly initialized model, all parameters are updated at the same rate; b) In the contrainstively pretrained model, $\theta$ is fine-tuned at a lower learning rate than $\phi$.}
\label{fig:learning_rate}
\end{figure}

\paragraph{Encoder Fine-Tuning for Classification Tasks}
In the training of the contrastively pretrained classification model, we allow fine-tuning of the learnable parameters $\theta$ in the encoder for the classification objective. In practice, the learning rate of $\theta$ is set to 5\,\% of than $\phi$, because the pretraining of the encoder already initializes $\theta$ at a relatively optimized position. In the randomly initialized model, the learning rate of all parameters is set to be the same as $\phi$ in the former group. Fig.~\ref{fig:learning_rate} shows the evolution of all tunable angles during the training of $N_L = 16$ on the quantum hardware.

\begin{figure}[h]
\centering
\includegraphics[width=0.8\textwidth]{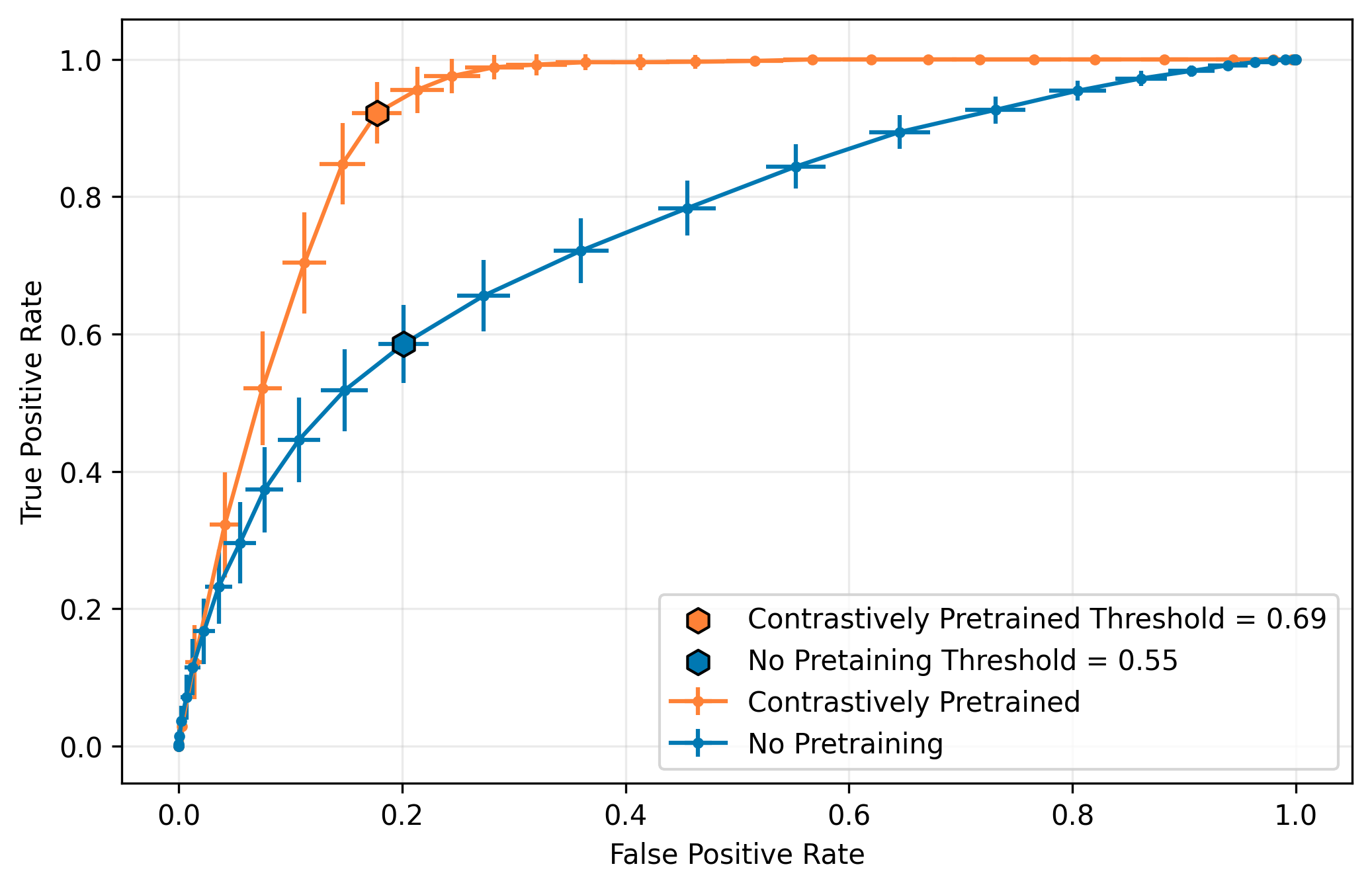}
\caption{\textbf{Receiver operating characteristic curve of simulated classifier models}. We sample 500 different training dataset with 12 training examples and perform the classifier training using noiseless simulation. The receiver operating curve on the aggregation of the 500 classifier tests are shown. The contrastively pretrained model outperforms the model without pretraining, but the best separation threshold is at 0.69 instead of 0.5. The average threshold that best separates the classes in the randomly initialized group is 0.55. }
\label{fig:ROC}
\end{figure}

\paragraph{Classification Threshold}
In Fig.~\ref{fig:results}(b), we show classifier accuracy when the classification threshold is fixed at 0.5. Another look into the receiver operating characteristics (ROC) of the classifiers reveals further insights into the behavior of these models. In many cases in the noiseless simulation results, we observe that the contrastively pretrained group provides good separation between two classes, but at a threshold different from 0.5, while the group without pretraining tends to separate relatively well around 0.5, though at a lower accuracy. As shown in Fig.~\ref{fig:ROC}, for example, with 12 training examples, the contrastively pretrained group outperforms the randomly initialized model as its ROC covers significantly more area under the curve. The separation point selected by the Youden's index is 0.69 for the former and 0.55 for the latter. 

We conjecture that this indicates a lack of expressiveness in $V_\phi$. If given more degrees of freedom, $V_\phi$ should be able to rotate the quantum features in a way that the projections of the two classes onto the final measurement basis produce a split at 0.5. The randomly initialized model suffers less from this problem because it does not have a limitation on the learning rate, allowing it to leverage the expressiveness in $U_\theta$ to achieve this end. The accuracy of the contrastively pretrained model can be further improved with more tunable parameters in $V_\phi$.

\bibliography{bibliography}

\end{document}